# Evolutionary Multitasking AUC Optimization

Chao Wang, Kai Wu, and Jing Liu, Xidian University, CHINA

*Abstract*—**Learning to optimize the area under the receiver operating characteristics curve (AUC) performance for imbalanced data has attracted much attention in recent years. Although there have been several methods of AUC optimization, scaling up AUC optimization is still an open due to its pairwise learning style. Maximizing AUC in the large-scale dataset can be considered as a non-convex and expensive problem. Inspired by the characteristic of pairwise learning, the cheap AUC optimization task with a small-scale dataset sampled from the large-scale dataset is constructed to promote the AUC accuracy of the original, large-scale, and expensive AUC optimization task. This paper develops an evolutionary multitasking framework (termed EMTAUC) to make full use of information among the constructed cheap and expensive tasks to obtain higher performance. In EMTAUC, one mission is to optimize AUC from the sampled dataset, and the other is to maximize AUC from the original dataset. Moreover, due to the cheap task containing limited knowledge, a strategy for dynamically adjusting the data structure of inexpensive tasks is proposed to introduce more knowledge into the multitasking AUC optimization environment. The performance of the proposed method is evaluated on a series of binary classification datasets. The experimental results demonstrate that EMTAUC is highly competitive to single task methods and online methods. Supplementary materials and source code implementation of EMTAUC can be accessed at https://github.com/xiaofangxd/EMTAUC.**

*Index Terms*—**AUC optimization, evolutionary multitasking optimization, expensive optimization, knowledge transfer, evolutionary machine learning.**

## I. INTRODUCTION

AUC (Area Under ROC Curve) [1] is an essential measure for characterizing the performance of machine learning methods in many real-world applications, such as ranking and anomaly detection tasks, and has become one of the most widely-used performance metrics for coping with imbalanced data. AUC maximization aims to obtain a real-valued ranking hypothesis that places a randomly drawn favorite instance with a higher decision value than a randomly drawn annoying instance. Many efforts have been devoted to designing efficient batch or online AUC optimization methods [2], [4], [9], [14].

For batch learning, Joachims [7] firstly developed a general framework for optimizing multivariate nonlinear performance measures such as the AUC and F1. Linear RankSVM [8] are the most commonly used AUC maximization methods. In addition, several efforts optimized the AUC maximization problem over mini-batches with practical strategies, such as U-statistics [9], [10], and the adaptive stochastic gradient method [12]. Unlike the above methods, there are several evolutionary algorithms (EA) based methods to handle this non-convex problem. The above methods optimize the convex surrogate of the AUC optimization problem. In general, EA-based methods transform the AUC optimization problem into a bi-objective optimization problem, which optimizes the True Positive Rate (TPR) and False Positive Rate (FPR) simultaneously. Several efforts [14] employed the multiobjective genetic programming (MOGP) algorithm for maximizing AUC, which regarded the problem of AUC maximization as a ROC convex hull (ROCCH) maximization problem. Recently, Zhao et al. [17], [18] considered a third objective, complexity, and proposed two novel 3D-convex-hull-based multiobjective EAs (MOEAs) for AUC maximization. Because the convex hull is not the same as the Pareto front, Qiu et al. [20] proposed a novel multi-level knee point-based EA (MKnEA-AUC) to overcome this limitation. However, current EA-based methods cannot cope with large-scale datasets due to their expensive cost. Many works employed online learning to address this limitation, which can be divided into two groups. The first group used a fixed-size buffer to store several sampled instances with various labels [21], and then the pairwise loss functions are calculated on this buffer. The second group [2], [4] kept only the first and second statistics for each instance and optimized the AUC metric in one pass through the data.

As mentioned above, batch learning methods are preferable to online AUC optimization methods if more attention is paid to performance. However, the high computational cost in evaluation makes it challenging to apply batch AUC optimization methods to large-scale datasets. Moreover, online AUC optimization methods have relatively low accuracy but can handle large-scale datasets due to the weak representation ability of the whole dataset. AUC optimization is an expensive problem, which demands the strategy to balance convergence and computational complexity. AUC optimization is related to evolutionary machine learning [49], [50]. When meeting large-scale data, Franco et al. [53], [54] introduced the integration of the GPU-based evaluation with the ILAS windowing scheme [51], [52]. Moreover, Franco et al. [50] introduced an automatic tuning strategy for rule-based evolutionary machine learning by using problem structure identification, especially for finding the

Corresponding Author: Kai Wu (Email: kwu@xidian.edu.cn).

Chao Wang, Kai Wu, and Jing Liu are with Xidian University, CHINA (e-mail: xiaofengxd@126.com; kwu@xidian.edu.cn; neouma@mail.xidian.edu.cn).

adequate set of hyperparameter values that is a very expensive process. Consequently, none of the current AUC methods attempts to develop cheap tasks to aid the performance of the expensive problem. None of them considers knowledge transfer among cheap and expensive tasks, which may promote AUC accuracy. The references most relevant to this idea are the multi-task Bayesian optimization [55] and evolutionary machine learning with minions [56], which used several tasks with small data to help quickly optimize the original task with big data. However, these methods do not consider the characteristics of AUC optimization with highlights in positive and negative instances. In this paper, two issues need to be addressed: 1) how to design the cheap task of the AUC optimization problem, and 2) how to transfer adequate knowledge between the cheap task and the expensive AUC optimization problem.

This paper develops an evolutionary multitasking AUC optimization framework to address two issues, termed EMTAUC. Due to the pair learning characteristics of AUC optimization, the function evaluation requires more time with the increasing number of instances. Thus, a small-scale dataset by sampling from the original dataset is established to solve the first issue. In this way, less time is needed to evaluate the AUC in the designed cheap task with the small-scale dataset. Moreover, due to the fast convergence of the cheap task, dynamically adjusting the data structure of inexpensive tasks is proposed to increase the diversity of knowledge contained in cheap tasks. Compared with the original task, the designed inexpensive task has partial knowledge because it contains part of the dataset. Thus, it is necessary to bridge the gap between the original task and the cheap task by dynamically adjusting the items of the small-scale dataset in the cheap task. In this strategy, the items with low AUC scores are filled into the sampled dataset. The solution of the second issue is inspired by evolutionary multitasking optimization (EMTO) [11], [25], a new paradigm for solving multiple optimization tasks by taking advantage of the parallelism mechanism of the human brain and EA. Recently, EMTO has been successfully applied to overcome many practical challenges [25]-[45] owing to its powerful search capability and easy scalability. These studies show that transferring valuable knowledge across tasks can improve the convergence of EAs. Unlike the existing surrogated-assisted EAs [46], [47], evolutionary multitasking optimization (EMTO) is adopted to use the knowledge of the designed cheap task and the original task to improve the AUC accuracy of a single task. To share knowledge between these two tasks, a multitasking AUC optimization environment is first established, where the models obtained from the cheap task are transferred to improve the performance of the expensive task and the better models acquired from the expensive task are shared to enhance the convergence of the cheap task.

To validate the performance of EMTUAC, a series of experiments on real-world datasets are conducted. Moreover, four state-of-the-art EMTO are embedded in EMTAUC. As shown in the experimental results, while AUC optimization is a desirable goal on its own, joining the designed cheap task and the original has a significant effect. The attendance of an inexpensive task significantly increases the AUC accuracy compared with performing this task in isolation. Moreover, the systematic comparison with existing gradient-based methods shows that EMTAUC matches or exceeds all the other algorithms. The highlights of the proposed EMTAUC are summarized as follows:

1) An evolutionary multitasking framework is first proposed to handle the tasks of AUC optimization. In EMTAUC, unlike existing methods that address the expensive AUC optimization task alone, a multitasking AUC optimization environment is established, where the designed cheap task and the original task are optimized simultaneously to promote the AUC performance.

2) Due to the partial knowledge of the designed cheap task, a strategy is proposed to bridge the gap between the original task and the cheap task by dynamically adjusting the data structure of cheap tasks. In this way, EMTAUC can enhance the representation ability of cheap tasks and break the limitation of online learning and batch learning methods.

The remainder of this paper is organized as follows. The related work on AUC optimization is introduced in Section II. Section III presents the background of the EMTO algorithm. Section IV gives a brief introduction on AUC optimization. Then an introduction on the designed EMTAUC is given in Section V. Section VI presents the experimental results to demonstrate the effectiveness of EMTAUC. Finally, Section VII concludes the work in this paper.

## II. Related Work

The standard process to maximize the AUC is described as follows. First, a classifier is trained based on the training instances. Then, the predicted test scores are sorted to obtain the AUC performance. Most of these approaches can be divided into three groups.

**Batch learning methods**. The first group of methods is based on batch learning. For batch learning, Joachims [7] firstly developed a general framework for optimizing multivariate nonlinear performance measures such as the AUC and F1. Linear RankSVM [8] is the most commonly used AUC maximization method. In addition, Gultekin et al. [9] employed U-statistics [10] over sampling mini-batches of positive/negative instance pairs to optimize the AUC maximization problem. Cheng et al. [12] proposed an adaptive stochastic gradient method over mini-batches, which employed a projection gradient strategy in its inner optimization. In these algorithms, special convex loss functions are designed. These algorithms are computationally challenging, and their computational costs are usually proportional to the number of positive-negative instance pairs.

**EA-based methods.** The second group employs EAs to improve AUC accuracy. The basic idea of EA-based methods is to transform the AUC optimization problem into a bi-objective optimization problem, where the True Positive Rate (TPR) and False Positive Rate (FPR) are employed as the two objectives. Gräning et al. [13] proposed a Pareto-based multiobjective EA to optimize TPR and FPR simultaneously. Wang et al. [14] proposed a multiobjective genetic programming (MOGP) algorithm for maximizing AUC. MOGP regarded AUC maximization as a ROC convex hull (ROCCH) maximization problem, of which the target is to maximize TPR and minimize FPR simultaneously. Due to the superiority of MOGP, several other MOEAs for maximizing AUC have been proposed. Wang et al. [15] further developed a convex hull-based MOGP. In this method, two new sorting and selection operations were designed to provide better solutions than MOGP. Hong et al. [16] proposed a new multiobjective EA for maximizing the ROCCH of neural networks (ETriCM), in which convex hull-based sorting with the convex hull of individual minima and an extreme area extraction selection was designed to avoid a local optimum. Recently, Zhao et al. [17], [18] proposed two novel 3D-convex-hull-based MOEAs for AUC maximization by considering three objectives, which have obtained good performance in application-oriented benchmarks. Moreover, the performance of the algorithms mentioned above greatly depends on the convex hull in the ROC space, but the convex hull is not precisely the same as the Pareto front. Qiu et al. [20] proposed a novel multi-level knee point-based EA to address this limitation. In addition, unlike the above methods, which optimize the AUC metric, Cheng et al. [19] proposed a multiobjective EA to optimize the partial AUC metric by designing a metric considering the partial range of the FPR.

**Online learning methods**. The third group investigates online learning methods for AUC optimization involving large-scale applications. Among the online AUC optimization methods, there are two core online AUC optimization frameworks. The first framework is based on the idea of buffer sampling [2], [21]. This framework used a fixed-size buffer to stand for the observed data for calculating the pairwise loss functions. For example, Zhao et al. [21] leveraged the reservoir sampling strategy to represent the empirical data by a fixed-size buffer, and then the pairwise loss functions are calculated on the fixed-size buffer. Kar et al. [2] introduced stream subsampling with replacement as the buffer update strategy. This strategy can improve the generalization capability of online learning algorithms for pairwise loss functions with buffer sampling.

The second framework takes a different perspective [3]. These methods extended the previous online AUC optimization framework with a regression-based one-pass learning mode, scanned through the training data. Because of the theoretical consistency between square loss and AUC, this framework achieved solid regret bounds by employing square loss for the AUC optimization task. For example, Gao et al. [3] exploited second-order statistics to identify the importance of less frequently occurring features and update their weights with a higher learning rate. Moreover, due to the failure in using the geometry knowledge of the data observed in the online learning process, Ding et al. [4] proposed an adaptive online AUC maximization by exploiting the knowledge of historical gradients to perform more informative online learning.

Although batch learning methods are preferable to online AUC optimization methods in terms of performance, they cannot handle large-scale datasets due to the high computational costs in evaluation. Moreover, online AUC optimization methods have relatively low accuracy but can handle large-scale datasets. Thus, it is necessary to design a strategy to balance convergence and computational complexity. Unlike current methods, a multitasking AUC optimization framework is proposed by designing a cheap task with a dynamic adjustment strategy to address the above issues. Due to the pair learning characteristics of AUC optimization, a cheap task is developed by sampling mini-batches of positive/negative instance pairs to avoid the limitation of batch AUC optimization methods. Moreover, due to the fast convergence of the cheap task and the local knowledge of the sampling dataset, dynamically adjusting the data structure of the cheap task is designed to make full use of the knowledge carried by the whole data. Then, any general EMTO algorithm can be employed to optimize the cheap and expensive tasks.

## III. EMTO

This section gives background knowledge on the multitasking optimization (MTO) problem first. Then, an overview of existing state-of-the-art methods for EMTO is also introduced, including multifactorial EA with online transfer parameter estimation (MFEAII) [28], symbiosis in biocoenosis optimization (SBO) [27], and evolutionary multitasking via explicit autoencoding (EMEA) [26].

### A. MTO Problem

MTO is an emerging topic for solving multiple tasks simultaneously with the benefits of implicit parallelism of EAs. Generally, for the $K$ minimization tasks, the MTO problem can be expressed as follows:

$$\min_{X_i} f_i(X_i), i = 1, 2, ..., K$$
$$s.t. X_i = [x_i^1, x_i^2, ..., x_i^{m_i}] \in D_i, i = 1, 2, ..., K \qquad (1)$$

where $f_i(X_i)$ represents the objective function of task $i$, and $X_i$ is the decision variable. $m_i$ is the dimension of $X_i$, and $D_i$ is the decision space for task $i$. MTO aims to find a set of solutions $X_i^* = \text{argmin } f_i(X_i)$ in the vast decision space by effective knowledge transfer across tasks. Generally speaking, EMTO algorithms need to consider two issues: 1. How to ensure knowledge flow in different decision spaces? 2. How to transfer the common knowledge across tasks? For the former, a unified search space $X$ of dimensionality $D = \max\{D_i\}$ in the multifactorial EA (MFEA) [25] is designed for all decision variables to ensure the flow of knowledge and then the key value of each variable is between 0 and 1. Unlike the establishment of a unified search space $X$ in MFEA, each task maintains its own search space's independence in EMEA [26]. The current mainstream EMTO algorithms can be divided into implicit and explicit paradigms. In the implicit paradigm, the common knowledge across tasks is transferred by genetic operators. In the explicit paradigm, the transferred solutions, including useful knowledge, are shared explicitly.

### B. MFEAII

MFEAII is the second generation of MFEA, the most popular MTO algorithm, which uses probabilistic models to effectively adjust the degree of knowledge transfer between tasks. In MFEAII, to provide a platform for knowledge sharing, the individual in a population is encoded in a unified search space introduced in Section III.A. Each task is viewed as a factor of the individual. Some concepts related to this factor are as follows.

1) *Factorial Cost $f_j^i$*: $f_j^i$ is the objective value of individual $i$ on task $j$.

2) *Factorial Rank $r_j^i$*: $r_j^i$ is the rank index of individual $i$ on the ascending factor cost list corresponding to task $j$.

3) *Skill Factor $\tau_i$*: $\tau_i$=argmin$\{r_j^i\}$ indicates the task associated with individual $i$.

4) *Scalar Fitness $\varphi_i$*: $\varphi_i = 1 / r_{\tau_i}^i$ is the scaled fitness of individual $i$.

In the evolutionary process of MFEAII, according to the skill factor, individuals in a population are implicitly divided into $K$ groups dedicated to $K$ tasks. Two genetic mechanisms, i.e., assortative mating and selective imitation, are designed to ensure that knowledge is transferred between tasks. More details of MFEAII can be found in [28].

### C. SBO

SBO is a novel paradigm for solving MTO problems, which leverages symbiosis in biocoenosis (SB) to transfer useful information across tasks. If a species has a positive/no positive/negative effect on other species, it is said that the species is beneficial/neutral/harmful to other species. According to the relationships between two species, SB is divided into six main types, including mutualism, commensalism, parasitism, neutralism, amensalism, and competition. The SB can reasonably simulate the information transmission in EMTO. The populations of various tasks form multiple species, and the information transmission across tasks can be regarded as symbiosis. The main framework of SBO holds three main components: 1) individual replacement strategy across paired tasks, 2) measurement of symbiosis by inter-task paired evaluations, and 3) dynamic adjustment strategy for the frequency and quantity of knowledge transfer based on SB. More details of SBO can be found in [27].

### D. EMEA

An explicit EMTO algorithm (called EMEA in this paper) was first proposed by Feng et al. [26], in which evolutionary solvers with different biases are used to solve different tasks simultaneously, and the common knowledge across tasks is transferred based on autoencoding. In this method, to make use of biases of multiple search operators, each task is assigned an independent solver. For any two tasks, a single-layer denoising autoencoder $M$ establishes the connection across tasks, which is trained by two sorted sets sampled from search spaces corresponding to the two tasks. Then multiple top individuals on each task are transferred to other tasks through the learned $M$. Next, populations with shared individuals undergo environmental selection until a specific stopping criterion is met. Thanks to the EA's implicit parallelism in EMEA, individuals are regarded as explicit knowledge transfer carriers. More details of EMEA can be found in [26].

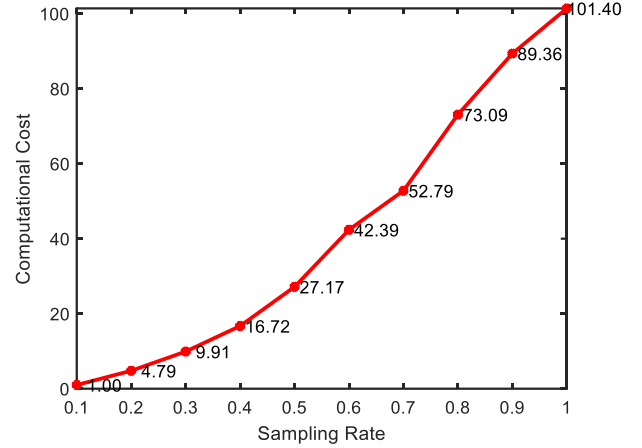

Figure 1 Variation of the average computational cost with the sampling rate on the a9a dataset.

## IV. AUC OPTIMIZATION

The AUC, an important metric, is widely employed to measure classification performance. The AUC optimization aims to minimize the pairwise loss between two instances from a training dataset. Let $(\mathbf{x}_n, y_n)$ be a training instance, where $\mathbf{x}_n \in \mathbb{R}^d$ and $y_n \in \{-1, +1\}$. Then a binary training set $D$ with $T^+$ positive instances and $T^-$ negative instances can be described as

$$S=\left\{\left(\mathbf{x}_1^+, +1\right),...,\left(\mathbf{x}_{T^+}^+, +1\right),\left(\mathbf{x}_1^-, -1\right),...,\left(\mathbf{x}_{T^-}^-, -1\right)\right\} \quad (2)$$

The objective function of AUC optimization on training set $D$ [21] is defined as

$$\text{AUC}(\mathbf{w}) = \frac{\sum_{i=1}^{T^+}\sum_{j=1}^{T^-} g_{\left(f(\mathbf{w}, \mathbf{x}_i^+) > f(\mathbf{w}, \mathbf{x}_j^-)\right)}}{T^+T^-} \quad (3)$$

where $g(\cdot)$ is the indicator function which outputs 1 if the argument is valid and 0 otherwise, and $f$: $\mathbb{R}^d \rightarrow \mathbb{R}$ is a real-valued function [3]. This paper focuses on learning a linear classification model $f = \mathbf{w}^T \cdot \mathbf{x}$. Then (3) can be rewritten as

$$
\begin{aligned}
\text{AUC}(\mathbf{w}) &= \frac{\sum_{i=1}^{T^+}\sum_{j=1}^{T^-} g_{\left(\mathbf{w}^T \cdot \mathbf{x}_i^+ > \mathbf{w}^T \cdot \mathbf{x}_j^-\right)}}{T^+T^-} \\
&= \frac{T^+T^- - \sum_{i=1}^{T^+}\sum_{j=1}^{T^-} g_{\left(\mathbf{w}^T \cdot \mathbf{x}_i^+ \leq \mathbf{w}^T \cdot \mathbf{x}_j^-\right)}}{T^+T^-} \\
&= 1 - \frac{\sum_{i=1}^{T^+}\sum_{j=1}^{T^-} g_{\left(\mathbf{w}^T \cdot \mathbf{x}_i^+ \leq \mathbf{w}^T \cdot \mathbf{x}_j^-\right)}}{T^+T^-}
\end{aligned}
\quad (4)
$$

TABLE I
THE COMPUTATIONAL COST OF THE AUC OPTIMIZATION ON THE A9A DATASET.

| Sampling Rate $s$ | Actual Computational Cost | Theoretical Computational Cost |
|---|---|---|
| 0.1 | 1 | $0.01 \times T^+ \times T^-$ |
| 0.2 | 4.788 | $0.04 \times T^+ \times T^-$ |
| 0.3 | 9.907 | $0.09 \times T^+ \times T^-$ |
| 0.4 | 16.721 | $0.16 \times T^+ \times T^-$ |
| 0.5 | 27.167 | $0.25 \times T^+ \times T^-$ |
| 0.6 | 42.394 | $0.36 \times T^+ \times T^-$ |
| 0.7 | 52.788 | $0.49 \times T^+ \times T^-$ |
| 0.8 | 73.086 | $0.64 \times T^+ \times T^-$ |
| 0.9 | 89.361 | $0.81 \times T^+ \times T^-$ |
| 1.0 | 101.401 | $T^+ \times T^-$ |

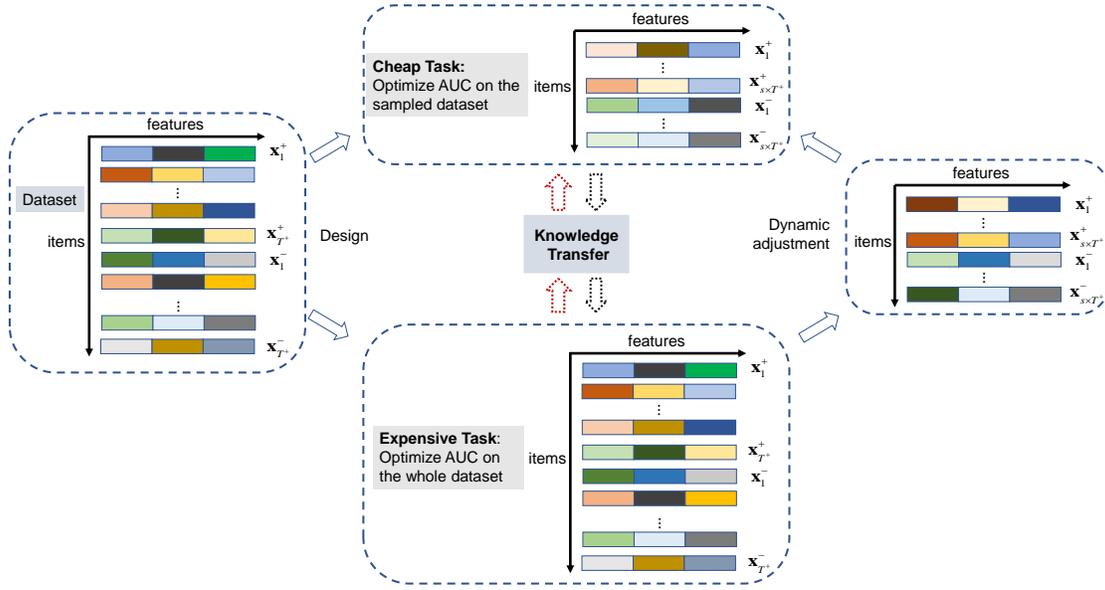

Figure 2 Construction of a multitasking AUC optimization environment.

It can be seen from the above equation that the AUC optimization is an expensive NP-hard problem. To find the optimal parameter of the linear classifier, the following objective is minimized

$$\min_{\mathbf{w}} \text{AUC}(\mathbf{w}) = \frac{\sum_{i=1}^{T^+}\sum_{j=1}^{T^-} g_{\left(\mathbf{w}^T \cdot \mathbf{x}_i^+ \leq \mathbf{w}^T \cdot \mathbf{x}_j^-\right)}}{T^+T^-} + \frac{\lambda}{2}|\mathbf{w}|^2 \quad (5)$$

where $|\mathbf{w}|^2$ is regularization. $\lambda$ is the penalty parameter, which balances the feasibility and sparsity of the model $f$. Note that different numbers of instances in the AUC optimization result in different computational costs. From (5), it can be seen that the time complexity of calculating the objective function of AUC optimization once is $O(T^+ \times T^-)$. As the number of instances increases, the optimization objective (5) becomes very expensive. For this reason, the a9a dataset [5] is used as an example to simulate the computational time of the objective function value with different sampling rates $s$ ranging from 0.1 to 1 over 1000 independent runs, where the sampling rate refers to the ratio of the number of instances used for training to the number of instances in the total dataset, and the proportion of positive and negative samples remains unchanged during the sampling process. Figure 1 and Table I show the variation of the average computational cost with the sampling rate on the a9a dataset, where the computational cost is obtained by normalizing the computational time of different sampling rates w.r.t. sampling rate 0.1. It can be seen from these results that the computational cost of each simulation increases with the number of instances in a flat manner, which is consistent with the theoretical computation time. Therefore, AUC optimization can be regarded as an expensive optimization problem when the dataset is large.

## V. EMTAUC

This section describes the detail of our proposed EMTAUC. Specifically, the cheap task's construction method is first given, providing a strong guarantee for the multitasking environment. Next, the knowledge available between the constructed cheap task and the expensive task is discussed. Then, dynamic adjustment strategies of the cheap task are presented. Finally, the framework of EMTAUC is introduced.

### A. Construction of a Multitasking AUC Optimization Environment

EMTO is one of the emerging areas of the computational intelligence community, which shows that optimizing multiple tasks simultaneously with effective knowledge transfer helps improve the convergence characteristics of each task. Inspired by this idea, this section designs a multitasking AUC optimization environment that includes cheap tasks and expensive tasks, as shown in Fig. 2, to utilize the information interaction between tasks to accelerate the convergence of expensive optimization tasks. In a multitasking AUC optimization environment, optimizing AUC on the whole dataset is regarded as an expensive task ($AUC_E$) and optimizing AUC on the sampled dataset is considered as a cheap task ($AUC_C$). Therefore, the objective function of multi-task AUC optimization can be defined as follows:

$$
\begin{cases}
\min_{\mathbf{w}} \ \mathrm{AUC}_E\left(\mathbf{w};S\right) = \dfrac{\sum_{i=1}^{T^+}\sum_{j=1}^{T^-} g_{\left(\mathbf{w}^\top\cdot\mathbf{x}_i^+\leq\mathbf{w}^\top\cdot\mathbf{x}_j^-\right)}}{T^+T^-} + \dfrac{\lambda}{2}|\mathbf{w}|^2 \\[4mm]
\min_{\mathbf{w}} \ \mathrm{AUC}_C\left(\mathbf{w};S^{'}\right) = \dfrac{\sum_{i=1}^{s\times T^+}\sum_{j=1}^{s\times T^-} g_{\left(\mathbf{w}^\top\cdot\mathbf{x}_i^+\leq\mathbf{w}^\top\cdot\mathbf{x}_j^-\right)}}{s^2 T^+T^-} + \dfrac{\lambda}{2}|\mathbf{w}|^2
\end{cases}
\tag{6}
$$

where $S^{'}$ is a simple dataset sampled from $S$ with a sampling rate $s$ randomly, and the proportions of positive and negative instances of $S'$ and $S$ are the same. If the computational cost of $AUC_C$ is 1, then the computational cost of $AUC_E$ can be expressed as

$$
\mathrm{cost}_{\mathrm{AUC}_E} = \frac{T^+T^-}{s^2 T^+T^-} = \frac{1}{s^2}
\tag{7}
$$

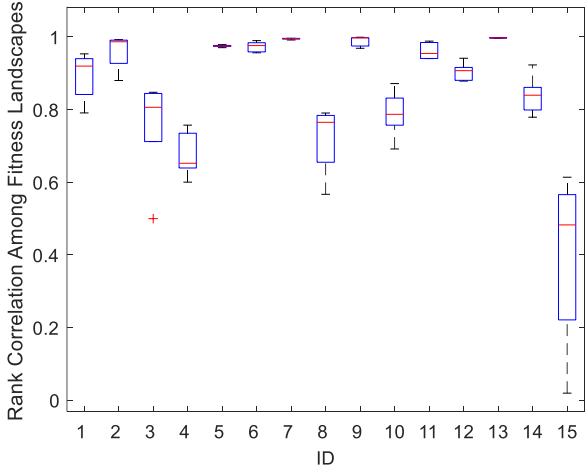

Figure 3 Rank (ordinal) correlation among fitness landscapes on 15 data sets under 50 runs.

In summary, by randomly sampling the dataset, a cheap task is constructed, which forms a multitasking AUC optimization environment with the original expensive task. Next, the knowledge available between the constructing cheap task and the expensive task is discussed to explain the feasibility of the constructed multitasking AUC optimization environment.

---

**Algorithm 1:** *Dynamic Adjustment Strategy of* $\mathrm{AUC}_C$

**Input:**
   $S$: the whole dataset used in $\mathrm{AUC}_E$;
   $\mathbf{w}_E$: the optimal classifier obtained on the current $\mathrm{AUC}_E$ task;
   $s$: the sampling rate;
**Output:**
   $S^{'}$: the adjusted dataset used in $\mathrm{AUC}_C$;

1.  $f^+, f^- \leftarrow \varnothing$;
2.  **for** $i = 1$ **to** $T^+$ **do**
3.     $f_i^+ \leftarrow \mathbf{w}_E^\top \mathbf{x}_i^+$;
4.     $f^+ \leftarrow f^+ \cup f_i^+$;
5.  **end for**
6.  **for** $i = 1$ **to** $T^-$ **do**
7.     $f_i^- \leftarrow \mathbf{w}_E^\top \mathbf{x}_i^-$;
8.     $f^- \leftarrow f^- \cup f_i^-$;
9.  **end for**
10. $score^+ = \{ \ score_i^+, 1 \leq i \leq T^+ \} \leftarrow 0, \ score^- = \{ \ score_i^-, 1 \leq i \leq T^- \} \leftarrow 0$;
11. **for** $i = 1$ **to** $T^+$ **do**
12.    **for** $j = 1$ **to** $T^-$ **do**
13.       **if** $(f_i^+ \leq f_j^-)$ **then**
14.          $score_i^+ \leftarrow score_i^+ + 1$;
15.       **end if**
16.    **end for**
17. **end for**
18. **for** $i = 1$ **to** $T^-$ **do**
19.    **for** $j = 1$ **to** $T^+$ **do**
20.       **if** $(f_i^- > f_j^+)$ **then**
21.          $score_i^- \leftarrow score_i^- + 1$;
22.       **end if**
23.    **end for**
24. **end for**
25. $S^{'} \leftarrow$ Select the top $sT^+$ positive instances and $sT^-$ negative instances from $S$ according to $score^+$ and $score^-$;

---

In MTO, the available knowledge across tasks can be measured by the overlap among the optimal solutions and the similarity among fitness landscapes [28]. For the constructed multitasking AUC optimization environment, all tasks have the same search space, that is, they all need to search for an optimal classifier $\mathbf{w}$. The similarity can be quantified by Spearman's rank correlation. Randomly generate 1,000,000 points in the search space $\mathbf{w}$, and then calculate their objective function values on the constructed cheap and expensive tasks to obtain the Spearman's rank (or the ordinal) correlation coefficient, which is considered as a representation of the similarity among fitness landscapes. The above steps are repeated 30 times. Each sampling process for constructing the cheap task is independent. The sample ratio $s$ is set to 10%. Figure 3 shows the rank (ordinal) correlation among fitness landscapes on 15 datasets under 50 runs. The datasets used are introduced in detail in Section VI.A. The average correlation coefficient is greater than 0.5 in most cases, which shows that the fitness landscapes of the cheap and expensive tasks are indeed similar. This phenomenon further illustrates that the constructed multitasking AUC optimization environment is feasible. In addition, on the 15th dataset, the variance of the rank (ordinal) correlation among fitness landscapes is extremely great, which shows that different sampling samples under a fixed sampling rate can greatly affect the similarity among fitness landscapes of the constructed cheap task and the expensive task. This is because cheap tasks composed of different sampling samples may contain different preference knowledge, resulting in different degrees of deviation between the data distribution of these cheap tasks and the expensive task.

---

**Algorithm 2:** *Framework of EMTAUC*

**Input**:
 $S$: the whole dataset used in $AUC_E$;
 $s$: the sampling rate;
 $\delta$: the generation interval for dynamic adjusting strategy;
 $\lambda$: the penalty parameter;
 $p$: the parameter of the MTO algorithm;
**Output**:
 $\mathbf{w}^*$: the optimal parameter of model $f$.
 1.  $F_1$, $F_2$←Initiate the $AUC_C$ and $AUC_E$ task according to (6);
 2.  $P_1$, $P_2$←Perform the **Initialization** module in an MTO algorithm (such as MFEAII, SBO, or EMEA);
 3.  $t$←1;
 4.  **while** (stopping criterion is not met) **do**
 5.    **if** mod($t$, $\delta$)=0 **then**
 6.      $F_1$←Perform the dynamic adjustment strategy of $AUC_C$ (Algorithm 1);
 7.      Re-evaluate $P_1$ on task $F_1$;
 8.    **end if**
 9.    $P_1$, $P_2$←Perform the **Main Loop** module in an MTO algorithm (such as MFEAII, SBO, or EMEA);
 10.   $t$←$t$+1;
 11.  **end while**
 12.  $\mathbf{w}^*$←Obtain the optimal parameter of model $f$ from $P_2$.

---

## B. Dynamic Adjustment Strategy of $AUC_C$

Compared with the optimization of the $AUC_E$ task, the $AUC_C$ task's optimization has a less difficult and computational cost, but the knowledge carried by the simple sampling dataset in $AUC_C$ is limited. For example, as shown in Fig. 3, in problems 3, 4, 8, 10, and 15, the average value of rank (ordinal) correlations among fitness landscapes are not very high. Therefore, this section proposes a dynamic adjustment strategy of $AUC_C$ to make full use of the knowledge in different sampling datasets, as shown in Algorithm 1. Assuming that the optimal classifier obtained on the current $AUC_E$ task is $\mathbf{w}_E$, the training instances used by the $AUC_C$ task are dynamically adjusted according to the performance of all data on the classifier $\mathbf{w}_E$. First, calculate all the instances' output values in the whole dataset on the classifier $\mathbf{w}_E$. Then assign a score to each instance to select those instances that are not easily identifiable by the classifier $\mathbf{w}_E$. The score of each positive instance is the number of negative instances whose output value is greater than or equal to the output value of the positive instance. Similarly, the score of each negative instance is the number of positive instances whose output value is less than the output value of the negative example. The greater the score of an instance, the more difficult it is to identify this instance. Therefore, the instances with high scores are employed to adjust the training data used by the $AUC_C$ task to make full use of the knowledge contained in these instances.

## C. Framework of EMTAUC

The framework of EMTAUC is shown in Algorithm 2. As can be observed, $AUC_C$ and $AUC_E$ tasks are initialized to build a multitasking AUC optimization environment. Then the two tasks are optimized simultaneously through any general EMTO algorithm such as MFEAII, SBO, and EMEA, where effective knowledge transfer between $AUC_C$ and $AUC_E$ tasks can speed up the convergence of expensive tasks. In addition, to ensure the diversity of knowledge transfer, the sampling dataset of $AUC_C$ tasks is dynamically adjusted after a certain number of generations ($\delta$ generations). According to the outputs of all samples on the optimal classifier obtained by the current $AUC_E$ task, the instances that are more difficult to identify are added to the $AUC_C$ tasks (see Algorithm 1). In this method, the maximum computational cost is the stopping criterion. The computational cost of a function evaluation of an $AUC_C$ task is considered as 1, and that of a function evaluation of an $AUC_E$ task is regarded as $1/s^2$.

TABLE II
DATASET STATISTICS.

| #ID | Dataset | $T^+ + T^-$ | $T^+$ | $T^-$ | $d$ |
|---|---|---|---|---|---|
| 1: | diabetes | 768 | 500 | 268 | 8 |
| 2: | fourclass | 862 | 307 | 555 | 2 |
| 3: | german | 1000 | 300 | 700 | 24 |
| 4: | splice | 1000 | 517 | 483 | 60 |
| 5: | usps | 7291 | 2596 | 4695 | 256 |
| 6: | australian | 690 | 307 | 383 | 14 |
| 7: | a9a | 32561 | 7841 | 24720 | 123 |
| 8: | sonar | 208 | 97 | 111 | 60 |
| 9: | svmguide1 | 3089 | 2000 | 1089 | 4 |
| 10: | svmguide3 | 1243 | 296 | 947 | 22 |
| 11: | segment | 2310 | 990 | 1320 | 19 |
| 12: | ijcnn1 | 49990 | 4853 | 45137 | 22 |
| 13: | satimage | 4435 | 1189 | 3246 | 36 |
| 14: | vowel | 528 | 240 | 288 | 10 |
| 15: | poker | 25010 | 23221 | 1789 | 10 |

## VI. EXPERIMENTS

### A. Experimental Setup

1) *Datasets*. In this paper, EMTAUC is evaluated on 15 datasets, including diabetes, fourclass, german, splice, usps, australian, a9a, sonar, svmguide1, svmguide3, segment, ijcnn1, satimage, vowel, and poker from the LIBSVM website [5]. These datasets are widely studied binary classification datasets. These datasets are commonly used for AUC optimization, as shown in many works [1]-[9]. The features for all datasets have been scaled to [-1,1], and the multi-class dataset is randomly divided into two groups, each of which contains the same number of classes. For each dataset, the number of negative and positive instances is shown in Table II, together with dimensionality $d$.

2) *Algorithm*. MFEAII [28], SBGA [27], EMEA [26], and SBCMAES [27], four state-of-the-art EMTO algorithms, are embedded in EMTAUC as the MTO optimizer to form EMTAUC-MFEAII, EMTAUC-SBGA, EMTAUC-EMEA, and EMTAUC-SBCMAES, respectively. In addition to the above EA-based methods, the nine state-of-the-art MOEAs or online AUC approaches are considered as baselines to demonstrate the effectiveness of EMTAUC in the experiments, which are ETriCM [16], MKnEA-AUC [20], OAM$_{seq}$, OAM$_{inf}$, OAM$_{gra}$ [21], Perceptron [22], CPA [23], CW [24], and PA [23]. ETriCM and MKnEA-AUC are representative methods based on convex hull and multi-level knee point, respectively, so they are selected as the compared algorithms. Moreover, two basic single-task solvers, the single-task genetic algorithm (Single-task GA) and CMAES [48], are also selected, which can independently optimize the expensive task. The performances of the compared methods are evaluated by five trials of 5-fold cross-validation, where the AUC values are obtained by averaging over these 50 runs, as summarized in Tables V, VI, and VII. Wins, ties, and losses of EMTAUC with respect to all other methods are reported in the last row of Tables VI and VII. If the evaluation of an algorithm cannot be obtained on each dataset, its results are set to "−". All experimental studies are performed on a Linux PC with Intel Core i7-10700K CPU at 3.80GHz and 32GB RAM.

TABLE III
THE COMPARISON OF EMTAUC AGAINST SINGLE TASK METHODS IN TERMS OF AUC.

| Dataset | EMTAUC-MFEAII | EMTAUC-SBGA | EMTAUC-EMEA | Single-task GA | EMTAUC-SBCMAES | CMAES |
|---|---|---|---|---|---|---|
| diabetes | **8.26E−01(3.34E−03)**+ | **8.21E−01(3.87E−03)**+ | **8.17E−01(5.50E−03)**+ | 8.08E−01(9.06E−03) | **8.35E−01(8.00E−04)**+ | 8.27E−01(5.00E−04) |
| fourclass | **8.34E−01(4.73E−04)**+ | 8.29E−01(2.84E−03)≈ | **8.30E−01(1.09E−03)**+ | 8.29E−01(4.65E−03) | **8.34E−01(8.90E−03)**+ | 8.28E−01(3.07E−04) |
| german_scale | **7.84E−01(1.76E−03)**+ | **7.71E−01(4.14E−03)**+ | **7.72E−01(6.05E−03)**+ | 7.61E−01(2.73E−03) | **7.92E−01(3.30E−03)**+ | 7.83E−01(2.60E−03) |
| splice | **8.19E−01(3.08E−03)**+ | **7.87E−01(8.48E−03)**+ | **7.82E−01(6.92E−03)**+ | 7.78E−01(7.45E−03) | **7.91E−01(2.76E−02)**+ | 7.62E−01(1.41E−02) |
| usps | **9.50E−01(1.53E−03)**+ | **9.37E−01(1.32E−03)**+ | **9.31E−01(3.01E−03)**+ | 9.29E−01(5.54E−03) | **5.29E−01(8.00E−03)**+ | 4.86E−01(1.90E−02) |
| australian | **9.22E−01(2.60E−03)**+ | **9.23E−01(1.82E−03)**+ | **9.21E−01(1.68E−03)**+ | 9.19E−01(1.02E−03) | 9.24E−01(2.80E−03)≈ | 9.28E−01(1.70E−03) |
| a9a | **8.90E−01(9.87E−04)**+ | **8.79E−01(1.68E−03)**+ | **8.72E−01(1.25E−03)**+ | 8.69E−01(6.24E−03) | **8.35E−01(6.12E−02)**+ | 5.47E−01(1.68E−02) |
| sonar | **8.44E−01(8.33E−03)**+ | **8.40E−01(7.13E−03)**+ | **8.39E−01(1.43E−02)**+ | 8.29E−01(1.09E−02) | **8.19E−01(2.40E−02)**+ | 7.76E−01(2.35E−02) |
| svmguide1 | **9.82E−01(4.41E−03)**+ | **9.83E−01(4.78E−03)**+ | **9.81E−01(3.65E−03)**+ | 9.77E−01(7.24E−03) | 9.89E−01(1.00E−04)≈ | 9.89E−01(1.00E−04) |
| svmguide3 | **7.42E−01(9.48E−03)**+ | **7.31E−01(1.12E−02)**+ | **7.27E−01(7.76E−03)**+ | 7.15E−01(4.64E−03) | **7.83E−01(3.40E−03)**+ | 7.76E−01(4.60E−03) |
| segment | **9.70E−01(2.77E−03)**+ | **9.66E−01(4.76E−03)**+ | **9.58E−01(5.90E−03)**+ | 9.54E−01(3.66E−03) | 9.85E−01(2.00E−03)≈ | 9.84E−01(1.40E−03) |
| ijcnn1 | **8.30E−01(9.62E−03)**+ | **7.75E−01(1.00E−02)**+ | **7.80E−01(8.59E−03)**+ | 7.41E−01(2.48E−03) | **8.85E−01(8.90E−03)**+ | 8.65E−01(1.86E−02) |
| satimage | **9.97E−01(8.26E−04)**+ | **9.97E−01(8.28E−04)**+ | **9.97E−01(6.23E−04)**+ | 9.96E−01(6.18E−04) | **9.99E−01(8.00E−04)**+ | 9.98E−01(1.60E−03) |
| vowel | **7.99E−01(3.65E−03)**+ | **8.01E−01(5.04E−03)**+ | **8.01E−01(6.33E−03)**+ | 7.87E−01(7.38E−03) | **8.00E−01(3.53E−03)**+ | 7.89E−01(4.60E−03) |
| poker | **5.21E−01(1.01E−03)**+ | **5.11E−01(3.03E−03)**+ | **5.11E−01(2.28E−03)**+ | 5.09E−01(2.87E−03) | **5.16E−01(4.00E−04)**+ | 5.06E−01(2.00E−04) |
| w/t/l to Single | 15/0/0 | 14/1/0 | 15/0/0 | – | 12/3/0 | – |

TABLE IV
THE COMPARISON OF EMTAUC AGAINST GRADIENT-BASED METHODS IN TERMS OF AUC.

| Dataset | $OAM_{seq}$ | $OAM_{gra}$ | $OAM_{inf}$ | CPA | CW | PA | Perceptron | EMTAUC |
|---|---|---|---|---|---|---|---|---|
| diabetes | 0.826(0.037)− | 0.826(0.034)− | – | – | – | – | 0.690(0.165)− | **0.833(0.002)** |
| fourclass | **0.831(0.025)** ≈ | **0.829(0.025)** ≈ | **0.831(0.020)** ≈ | 0.812(0.020)− | 0.758(0.032)− | 0.668(0.0168)− | 0.701(0.039)− | **0.832(0.0003)** |
| german | 0.775(0.041)− | 0.772(0.036)− | – | – | – | – | – | **0.796(0.001)** |
| splice | 0.859(0.019)− | **0.886(0.017)**+ | – | – | – | – | – | 0.878(0.001) |
| usps | 0.931(0.016)− | 0.935(0.012)− | – | – | – | – | – | **0.965(0.0007)** |
| a9a | 0.842(0.017)− | 0.857(0.017)− | – | – | – | – | – | **0.896(0.0001)** |
| sonar | 0.850(0.042)− | 0.849(0.043)− | 0.849(0.043)− | 0.827(0.052)− | 0.793(0.059)− | 0.790(0.057)− | 0.780(0.060)− | **0.856(0.010)** |
| svmguide1 | **0.988(0.003)** ≈ | **0.988(0.002)** ≈ | **0.989(0.002)** ≈ | 0.885(0.008)− | 0.922(0.019)− | 0.799(0.219)− | 0.883(0.140)− | **0.989(0.0001)** |
| svmguide3 | 0.760(0.035)− | 0.755(0.034)− | 0.768(0.035)− | 0.707(0.037)− | 0.723(0.036)− | 0.696(0.038)− | 0.696(0.038)− | **0.795(0.004)** |
| segment | 0.956(0.013)− | 0.955(0.014)− | 0.956(0.013)− | 0.886(0.021)− | 0.896(0.021)− | 0.852(0.024)− | 0.852(0.024)− | **0.987(0.0008)** |
| ijcnn1 | 0.921(0.008)− | 0.910(0.009)− | **0.928(0.015)** ≈ | 0.910(0.011)− | 0.829(0.021)− | 0.531(0.074)− | 0.647(0.088)− | **0.927(0.006)** |
| satimage | 0.919(0.014)− | 0.911(0.017)− | 0.921(0.013)− | 0.828(0.024)− | 0.619(0.024)− | 0.646(0.024) | 0.605(0.025)− | **0.999(0.00003)** |
| vowel | **0.931(0.046)**+ | 0.928(0.046)+ | 0.931(0.041)+ | 0.923(0.032)+ | 0.870(0.063)+ | 0.863(0.063)+ | 0.859(0.055)+ | 0.806(0.0001) |
| poker | **0.592(0.060)**+ | 0.586(0.062)+ | **0.594(0.067)**+ | 0.524(0.067)+ | 0.457(0.058)− | 0.506(0.023)− | 0.502(0.031)− | 0.517(0.0001) |
| w/t/l | 2/2/10 | 3/2/10 | 2/2/6 | 2/0/8 | 1/0/9 | 1/0/9 | 1/0/9 | – |

3) *Parameter Settings*: All the above genetic operator-based algorithms (ETriCM [16], MKnEA-AUC [20], MFEAII, EMEA, SBGA, and Single-task GA) employ the same SBX crossover and PM mutation for a fair comparison, where the distribution indexes of SBX and PM are set to 15 and 15, respectively. The probabilities of SBX and PM are set to 1.0 and $1/d$, respectively. Then the population size of each task in EMEA, SBGA, Single-task GA, SBCMAES, and CMAES is set to 10, and the population size in MFEAII is set to 20. For SBGA, beneficial factor $B$ and harmful factor $H$ are set to 0.25 and 0.5, respectively. For EMEA, the interval of individual transfer $G$ and the number of transferred individuals $S$ are set to 5 and 2, respectively. For MFEAII, the probability model is configured as a variablewise (marginal) normal distribution. To demonstrate the effectiveness of the proposed EMTAUC compared with the single-task method, $\delta = 30$, $\lambda = 2^{-3}$, and the sample ratio $s$ is set to 10%. The maximum computational cost is set to 101000. The computational cost of a function evaluation of the cheap task is considered as 1, and that of a function evaluation of the expensive task is regarded as $1/s^2$, where $s$ is the sampling rate.

Moreover, to compare with the MOEAs and online learning methods, the population size of each task in EMTAUC is set to 100, the ratio $s$ is set to 10%, and the number of transferred individuals $S$ is set to 5. All algorithms are stopped until the algorithms converge. For EMTAUC, the interval $\delta$ of the dynamic adjustment strategy is chosen from {5:5:40} with the best performance. All other parameters of the comparison algorithm are tuned for the current problems. Five-fold cross-validation is executed on training sets to decide the learning rate $l \in 2^{[-12:10]}$ for online algorithms and penalty parameter $\lambda \in 2^{[-10:10]}$ for all algorithms. For $OAM_{seq}$, $OAM_{inf}$, and $OAM_{gra}$, the buffer sizes are fixed to 100, as suggested in [21]. For ETriCM and MKnEA-AUC, the population size is set to 100. CHIM-based sorting and EAE-selection strategy are employed to choose individuals in ETriCM [16].

### B. Results and Discussion

This section first studies the effectiveness of the EMTAUC framework compared with the single task framework. The results are reported in Table III. The symbols "−", "≈" and "+" imply that the corresponding compared method is significantly worse, similar, and better than the Single-task GA or CMAES on the Wilcoxon rank-sum test with a 95% confidence level, respectively. On each of the 15 real-world datasets, the AUC of EMTAUC matches or exceeds that of the single-task method in all cases. EMTAUC-SBGA outperforms the single task in 13 of 14 cases and matches the single task on the fourclass dataset. EMTAUC-MFEAII and EMTAUC-EMEA are superior to Single-task GA in all cases. Compared to CMAES, EMTAUC-CMAES performs better on 12 and ties 3 out of 15 cases in terms of the AUC. In EMTAUC, practical knowledge is transferred between the cheap task and the expensive task to improve the performance of the AUC optimization problem. Since EMTAUC has the same basic evolutionary solver of the inexpensive task and expensive task as the single-task method, the effectiveness of knowledge transfer in EMTAUC is confirmed.

TABLE V
THE COMPARISON OF EMTAUC AGAINST MOEAs IN TERMS OF AUC.

| Dataset | ETriCM | MKnEA-AUC | EMTAUC |
|---|---|---|---|
| Australian | 0.907(0.745)− | 0.904(0.663)− | **0.927(0.0004)** |
| Transfusion | 0.786(0.713)− | 0.794(0.309)− | **0.832(0.0009)** |
| Sonarall | 0.887(1.385)+ | **0.912(0.954)+** | 0.856(0.010) |
| pima_ind ians_diabctes | 0.737(0.513)− | 0.750(0.401)− | **0.830(0.0004)** |
| Muskl | 0.904(0.844)− | 0.914(0.730)− | **0.921(0.0005)** |
| Musk2 | 0.958(0.229)≈ | 0.961(0.208)≈ | 0.964(0.0004) |
| Hill_Valley_without | **1.000(0.005)+** | **1.000(0.000)+** | 0.958(0.008) |
| Wla | 0.912(0.165)− | 0.913(0.190)− | **0.969(0.0002)** |
| Parkinsons | 0.776(2.029)− | 0.808(2.329)− | **0.892(0.004)** |
| Spcctf | 0.830(1.978)− | 0.857(0.843)− | **0.868(0.0008)** |
| Splice | 0.912(0.258)− | 0.919(0.174)− | **0.927(0.0001)** |
| Fourclass | 0.806(0.352)− | 0.808(0.156)− | **0.832(0.0003)** |
| w/t/l | 9/1/2 | 9/1/2 | − |

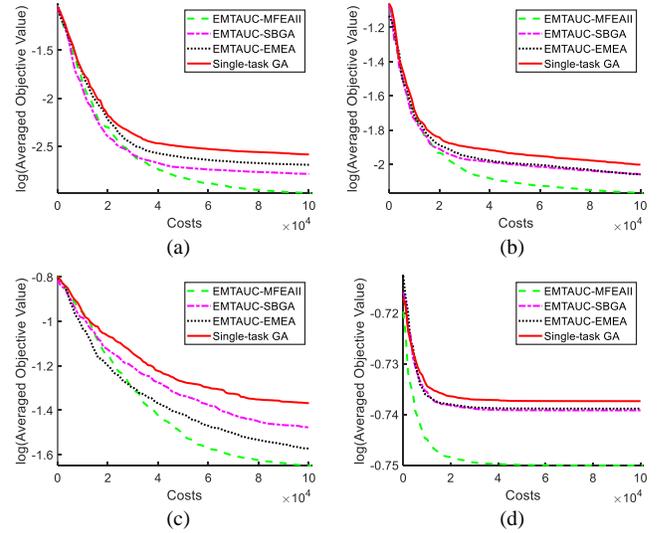

Figure 4 The convergence curve of EMTAUC-MFEAII, EMTAUC-SBGA, EMTAUC-EMEA, and Single-task GA on four representative datasets, (a) usps, (b) a9a, (c) ijcnn1, and (d) poker.

In addition, the average convergence trends of four compared methods, i.e., EMTAUC-MFEAII, EMTAUC-SBGA, EMTAUC-EMEA, and Single-task GA, on usps, a9a, ijcnn1, and poker datasets are shown in Fig. 4 to illustrate the effectiveness of our proposal further. Due to the space limitation, the average convergence trends of four compared methods on all datasets are shown in Fig. S2 (see Supplementary Material). In these figures, the x-axis represents the computational cost, and the y-axis represents the average objective value on a log scale. As can be seen from Fig. 4, EMTAUC-MFEAII, EMTAUC-SBGA, and EMTAUC-EMEA show excellent performance in the average convergence trend compared with Single-task GA in most cases, which illustrates that knowledge transfer across tasks can effectively accelerate the convergence of expensive tasks. In addition, EMTAUC-MFEA-II significantly outperforms the other multitasking methods, especially on the poker dataset. As shown in Fig. 3, the intertask Spearman rank correlation is not high on the poker dataset, so harmful knowledge across tasks may prevail in the complex multitasking AUC environment. MFEAII can theoretically minimize the trend of negative transfer across tasks, while other methods do not explicitly consider preventing the process, such as SBGA and EMEA. Therefore, EMTAUC-MFEA-II shows higher performance on most datasets.

The experiments are designed to compare EMTAUC with online learning methods, as shown in Table IV. The symbols "−", "≈" and "+" imply that the corresponding compared method is significantly worse, similar to, and better than EMTAUC on the Wilcoxon rank-sum test with a 95% confidence level, respectively. The AUC of EMTAUC outperforms CW, PA, and Perceptron in 9 of 10 cases and loses once to them on the vowel dataset. EMTAUC outperforms CPA in 8 out of 10 cases and loses two times on vowel and poker datasets. $\text{OAM}_{\text{inf}}$ exceeds EMTAUC in 2 of 10 cases, matches two times, and loses six times to EMTAUC. EMTAUC matches or exceeds $\text{OAM}_{\text{seq}}$ in 12 of 14 cases and loses two times to $\text{OAM}_{\text{seq}}$ on vowel and poker datasets. EMTAUC matches or exceeds $\text{OAM}_{\text{seq}}$ in 11 of 14 cases and loses three times to $\text{OAM}_{\text{gra}}$ on splice, vowel, and poker datasets. This phenomenon illustrates the advantages of EMTAUC.

TABLE VI
THE COMPARISON OF EMTAUC WITH AND WITHOUT THE DYNAMIC ADJUSTMENT STRATEGY OF $AUC_c$ (EMTAUC AND EMTAUC-I) IN TERMS OF AUC.

| Dataset | EMTAUC-MFEAII | EMTAUC-I-MFEAII | EMTAUC-SBGA | EMTAUC-I-SBGA | EMTAUC-EMEA | EMTAUC-I-EMEA |
|---|---|---|---|---|---|---|
| diabetes | 8.26E−01(3.34E−03)≈ | 8.21E−01(4.11E−03) | 8.21E−01(3.87E−03)≈ | 8.24E−01(4.38E−03) | 8.17E−01(5.50E−03)≈ | 8.19E−01(1.44E−02) |
| fourclass | **8.34E−01(4.73E−04)+** | 8.25E−01(1.25E−03) | 8.29E−01(2.84E−03)≈ | 8.29E−01(2.35E−03) | 8.30E−01(1.09E−03)≈ | 8.29E−01(2.12E−03) |
| german_scale | **7.84E−01(1.76E−03)+** | 7.78E−01(1.49E−03) | **7.71E−01(4.14E−03)+** | 7.67E−01(3.90E−03) | **7.72E−01(6.05E−03)+** | 7.67E−01(6.53E−03) |
| splice | **8.19E−01(3.08E−03)+** | 8.11E−01(3.40E−03) | **7.87E−01(8.48E−03)+** | 7.77E−01(3.67E−03) | **7.82E−01(6.92E−03)+** | 7.72E−01(3.58E−03) |
| usps | 9.50E−01(1.53E−03)≈ | 9.49E−01(1.14E−03) | **9.37E−01(1.32E−03)+** | 9.31E−01(2.61E−03) | 9.31E−01(3.01E−03)≈ | 9.31E−01(2.61E−03) |
| australian | 9.22E−01(2.60E−03)≈ | 9.25E−01(1.33E−03) | 9.23E−01(1.82E−03)≈ | 9.24E−01(1.62E−03) | 9.21E−01(1.68E−03)≈ | 9.23E−01(2.38E−03) |
| a9a | 8.90E−01(9.87E−04)≈ | 8.88E−01(6.57E−04) | 8.79E−01(1.68E−03)≈ | 8.78E−01(1.04E−03) | 8.72E−01(1.25E−03)≈ | 8.72E−01(2.44E−03) |
| sonar | **8.44E−01(8.33E−03)+** | 8.21E−01(1.03E−02) | **8.40E−01(7.13E−03)+** | 8.13E−01(1.30E−02) | **8.39E−01(1.43E−02)+** | 8.27E−01(1.08E−02) |
| svmguide1 | 9.82E−01(4.41E−03)≈ | 9.84E−01(3.22E−03) | 9.83E−01(4.78E−03)≈ | 9.86E−01(5.48E−03) | 9.81E−01(3.65E−03)≈ | 9.83E−01(6.16E−03) |
| svmguide3 | **7.42E−01(9.48E−03)+** | 7.33E−01(5.75E−03) | **7.31E−01(1.12E−02)+** | 7.20E−01(9.28E−03) | **7.27E−01(7.76E−03)+** | 7.14E−01(8.02E−03) |
| segment | 9.70E−01(2.77E−03)≈ | 9.72E−01(6.18E−03) | **9.66E−01(4.76E−03)+** | 9.56E−01(6.49E−03) | 9.58E−01(5.90E−03)≈ | 9.60E−01(4.77E−03) |
| ijcnn1 | **8.30E−01(9.62E−03)+** | 8.18E−01(8.02E−03) | 7.75E−01(1.00E−02)− | **7.91E−01(1.02E−02)** | **7.80E−01(8.59E−03)+** | 7.73E−01(7.34E−03) |
| satimage | 9.97E−01(8.26E−04)≈ | 9.97E−01(1.37E−03) | 9.97E−01(8.28E−04)≈ | 9.97E−01(7.80E−04) | 9.97E−01(6.23E−04)≈ | 9.97E−01(9.94E−04) |
| vowel | 7.99E−01(3.65E−03)≈ | 7.98E−01(3.85E−03) | **8.01E−01(5.04E−03)+** | 7.81E−01(3.10E−03) | **8.01E−01(6.33E−03)+** | 7.81E−01(5.61E−03) |
| poker | **5.21E−01(1.01E−01)+** | 5.07E−01(1.24E−03) | **5.11E−01(3.03E−03)+** | 5.08E−01(1.67E−03) | **5.11E−01(2.28E−03)+** | 5.08E−01(2.08E−03) |
| w/t/l | 7/8/0 | − | 8/6/1 | − | 7/8/0 | − |

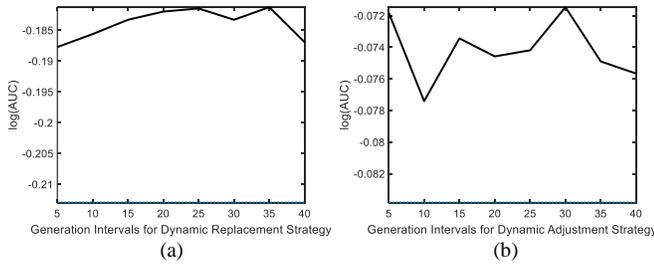

Figure 5 The relationships between the generation intervals for dynamic adjusting strategy and the average AUC value obtained by EMTAUC-SBGA on two representative datasets, where the dotted line represents the average AUC value obtained by single-task GA, (a) diabetes, and (b) australian.

In addition, Table V shows the comparison results of EMTAUC against MOEAs in terms of AUC. The datasets used in this experiment can be found from the LIBSVM and UCI web site [5]. EMTAUC outperforms ETriCM and MKnEA-AUC in 9 out of 12 cases, ties once, and loses two times, which further shows that our proposal has excellent performance.

### C. Effectiveness of Dynamic Adjustment Strategy of $AUC_C$

Our proposal establishes a multitasking AUC optimization environment to take advantage of the similarity across the constructed cheap task $\text{AUC}_c$ and the original task $\text{AUC}_E$. However, the knowledge carried by the task $\text{AUC}_c$ may be incomplete. Therefore, a dynamic adjustment strategy of $\text{AUC}_c$ is proposed to make full use of the knowledge carried by the whole datasets. In this section, the effectiveness of the dynamic adjustment strategy of $\text{AUC}_c$ is analyzed.

Table VI shows the AUC of the EMTAUC with and without the dynamic adjustment strategy of $\text{AUC}_c$ (EMTAUC and EMTAUC-I) on all datasets, where the basic MTO solvers in EMTAUC are respectively set as MFEAII, SBGA, and EMEA. EMTAUC can be observed to win EMTAUC-I on problems 3, 4, 8, 10, and 15 in terms of AUC values, which further illustrates the effectiveness of the dynamic adjustment strategy of $\text{AUC}_c$. As shown in Fig. 3, the average value of rank (ordinal) correlations among fitness landscapes in these problems are relatively low. By dynamically adjusting the sampled data in the $\text{AUC}_c$, new knowledge is introduced into the multitasking AUC environment to improve the performance of EMTAUC.

### D. Parameters Analysis

This section analyzes the effect of three critical parameters in EMTAUC and takes the algorithm EMTAUC-SBGA as an example. Three key parameters are summarized as follows: 1) $\delta$, the generation intervals for dynamic adjusting strategy; 2) $s$, the ratio of the whole dataset for the cheap task; 3) $\lambda$: the penalty parameter. The values of other parameters are fixed to analyze one parameter visually.

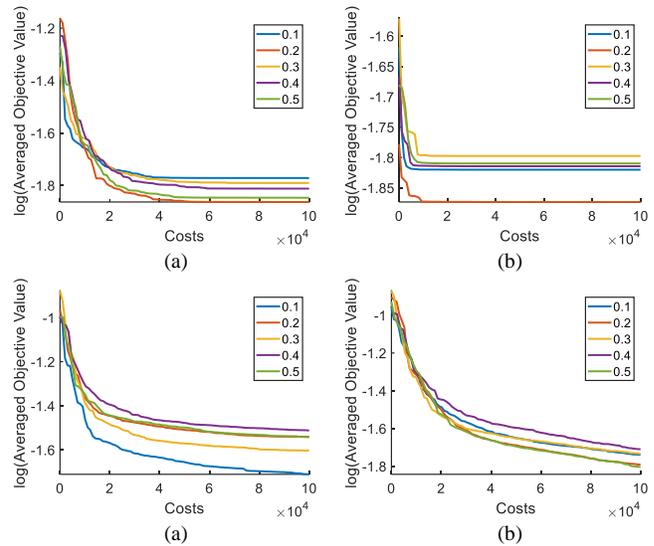

Figure 6 The performance of EMTAUC with varying ratio of the whole dataset for the cheap task on (a) diabetes, (b) fourclass, (c) german, and (d) splice.

Figure 5 shows the average AUC value of EMTAUC versus varying $\delta$ on diabetes and australian datasets. The average AUC value of EMTAUC varies with $\delta$ on all datasets as shown in Fig. S3 (see Supplementary Material). The value of $\delta$ is set from 5 to 40 in steps of 5. It can be seen that the value of $\delta$ has an essential effect on the performance of EMTAUC. The parameter affects the frequency of dynamic adjustment of the cheap task. Moreover, in Fig. 5(a), the value of $\delta$ should be set to 5 to obtain the best average objective value on the diabetes dataset. However, in Fig. 5(b), EMTAUC can obtain the best performance on the australian dataset if $\delta = 10$. Thus, the collection of the parameter on the diabetes dataset may not be suitable for the others. $\delta$ should be set to a relatively small value. In addition, the change curve is located above the straight line corresponding to the average AUC value obtained by Single-task GA in most cases, showing that no matter what the value of $\delta$ is, useful knowledge will be transferred across tasks to improve the convergence performance.

Figure 6 shows the average objective value of EMTAUC versus varying $s$ on diabetes, fourclass, german, and splice datasets. The average objective value of EMTAUC varies with $s$ on all datasets as shown in Fig. S4 (see Supplementary Material). In Fig. 6, the average objective value increases with increasing computational costs. Figure 6 (a) shows that the performance of EMTAUC on the diabetes dataset is the best in all cases when $s$ is set to 0.2. Figure 6 (b) shows that when $s$ is set to 0.3, the performance of EMTAUC is the best in all cases. With the increase of $s$, the AUC obtained by EMTAUC first increases and then decreases. This phenomenon can be found in Fig. S4 (a), (b), (d), (e), (f), (g), (h), (k), (l), (m), (n), and (o). This phenomenon appears because the relatively high ratio of the whole dataset may own the better representation of the original dataset, but a higher value of $s$ may break the balance of the computational costs and the convergence and result in the low AUC accuracy. In Fig. 6(c), the performance of EMTAUC is the best in all cases when $s$ is set to 0.1. With the increase of $s$, the performance of EMTAUC decreases. This phenomenon appears because a higher value of $s$ may break the balance of the computational costs and the convergence and result in low AUC accuracy. This case can also be found in Fig. S4(i) and Fig. S4(j).

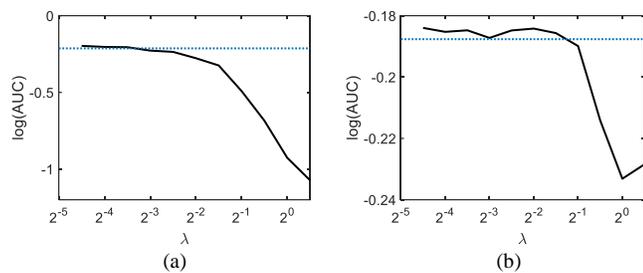

Figure 7 The relationships between the penalty parameter and the average AUC value obtained by EMTAUC-SBGA on two representative datasets, where the dotted line represents the average AUC value obtained by Single-task GA, (a) diabetes, and (b) fourclass.

Figure 7 shows the average AUC value of EMTAUC versus varying $\lambda$ on diabetes and fourclass datasets. The average AUC value of EMTAUC varies with $\lambda$ on all datasets as shown in Fig. S5 (see Supplementary Material). The value of $\lambda$ is set to $2^i$, $i=-5$, $-4$, $-3$, …, 1. $\lambda$ balances the feasibility and sparsity of model $f$. Figure 7 shows that the smaller value of $\lambda$ denotes the lower average AUC value. To achieve the high performance of EMTAUC, more attention should be paid to the first term of (5). Thus, $\lambda$ should be set to a value small than $2^{-3}$.

## VII. Conclusions

This paper proposes a multitasking AUC optimization framework to overcome the limitations of batch and online AUC optimization methods. First, an auxiliary task by sampling the mini-batches of the original dataset is designed. The other task is to optimize the original problem with the whole dataset. Moreover, a dynamic adjustment strategy is proposed to enhance the representation ability of the sampled dataset by replacing the items with the cases with low AUC scores in the original dataset. The results compared with the batch EA method have demonstrated the advantage of the proposed multitasking AUC optimization framework. Compared with online AUC optimization methods, EMTAUC is highly competitive in small-scale and large-scale datasets.

Our work raises many questions to further the development of the evolutionary multitasking AUC optimization problem. For example, the parameters have a significant effect on the performance of EMTAUC. In future work, we will develop an effective tuning-free EMTAUC with a policy network. In addition, applying EMTAUC to solve other machine learning problems is also a promising research topic. Due to its extensibility, EMTAUC has preliminarily been applied to tune hyperparameters of LightGBM [57] to further illustrate the advanced nature of our proposal, which can be found in the Supplementary Material. The experimental results conducted on 11 datasets have shown the advantages of our proposal over single task optimization.

## Acknowledgment

This work was supported in part by the Key Project of Science and Technology Innovation 2030 supported by the Ministry of Science and Technology of China under Grant 2018AAA0101302 and in part by the General Program of National Natural Science Foundation of China (NSFC) under Grant 61773300.